\documentclass[conference]{IEEEtran}

\usepackage{xcolor}
\definecolor{wong-black}        {HTML}{000000}
\definecolor{wong-lightorange}  {HTML}{E69F00}
\definecolor{wong-lightblue}    {HTML}{56B4E9}
\definecolor{wong-green}        {HTML}{009E73}
\definecolor{wong-yellow}       {HTML}{F0E442}
\definecolor{wong-darkblue}     {HTML}{0072B2}
\definecolor{wong-darkorange}   {HTML}{D55E00}
\definecolor{wong-pink}         {HTML}{CC79A7}

\usepackage{cite}
\usepackage{csquotes}
\usepackage{amsmath,amssymb,amsfonts}
\usepackage{algorithmic}
\usepackage{graphicx}
\usepackage{textcomp}
\usepackage{xcolor}
\usepackage[nolist, nohyperlinks, printonlyused]{acronym} 
\usepackage{booktabs}
\usepackage{pifont} 
\usepackage{url}

\usepackage{hyperref} 
\hypersetup{
    colorlinks=true,
    citecolor=wong-green,
    linkcolor=wong-darkblue,
    filecolor=wong-pink,      
    urlcolor=wong-black
}

\usepackage{tikz}
\usetikzlibrary{calc, intersections, positioning, arrows.meta, shadows}

\def\BibTeX{{\rm B\kern-.05em{\sc i\kern-.025em b}\kern-.08em
    T\kern-.1667em\lower.7ex\hbox{E}\kern-.125emX}}

\newcommand{\cmark}{\ding{51}}%

\graphicspath{{img/}}    
\begin{document}

\title{Exploring the Potential of World Models\\for Anomaly Detection in Autonomous Driving}

\author{\IEEEauthorblockN{Daniel Bogdoll\IEEEauthorrefmark{2}\IEEEauthorrefmark{3},
Lukas Bosch\IEEEauthorrefmark{3},
Tim Joseph\IEEEauthorrefmark{2},
Helen Gremmelmaier\IEEEauthorrefmark{2},
Yitian Yang\IEEEauthorrefmark{2},
and J. Marius Zöllner\IEEEauthorrefmark{2}\IEEEauthorrefmark{3}}

\IEEEauthorblockA{\IEEEauthorrefmark{2}FZI Research Center for Information Technology, Germany\\
bogdoll@fzi.de}
\IEEEauthorblockA{\IEEEauthorrefmark{3}Karlsruhe Institute of Technology, Germany\\}}

\maketitle

\begin{acronym}
    \acro{ml}[ML]{Machine Learning}
	\acro{cnn}[CNN]{Convolutional Neural Network}
	\acro{dl}[DL]{Deep Learning}
	\acro{ad}[AD]{Autonomous Driving}
        \acro{VAE}[VAE]{Variational Autoencoder}
        \acro{vrkn}[VRKN]{Variational Recurrent Kalman Networks}
        \acro{MLP} [MLP] {Multi-Layer Perceptron}
        \acro{GRU} [GRU] {Gated Recurrent Unit}
        \acro{PSNR} [PSNR] {Peak Signal to Noise Ratio}
        \acro{AE} [AE] {Autoencoder}
        \acro{CL-VAE} [CL-VAE] {Conditional Latent Space Variational Autoencoder}
        \acro{MILE} [MILE] {Model-Based Imitation Learning}
        \acro{IL} [IL] {Imitation Learning}
        \acro{RL} [RL] {Reinforcement Learning}
        \acro{POMDP} [POMDP] {Partially Observable Markov Decision Processes}
        \acro{BeV} [BeV] {Bird's-Eye View}
\end{acronym}

\begin{abstract}
In recent years there have been remarkable advancements in autonomous driving. While autonomous vehicles demonstrate high performance in closed-set conditions, they encounter difficulties when confronted with unexpected situations. 
At the same time, world models emerged in the field of model-based reinforcement learning as a way to enable agents to predict the future depending on potential actions. This led to outstanding results in sparse reward and complex control tasks. This work provides an overview of how world models can be leveraged to perform anomaly detection in the domain of autonomous driving. We provide a characterization of world models and relate individual components to previous works in anomaly detection to facilitate further research in the field.
\end{abstract}


\section{Introduction}
\label{sec:introduction}

The detection of anomalies, or corner cases, is a challenging task with applications in autonomous driving. Neural networks, used in tasks like semantic segmentation, tend to be overconfident when confronted with the unseen. Therefore, an anomaly detection system can increase the reliability of autonomous systems which build on such components~\cite{termohlenCornerCaseDetection2019}.
In perception systems of autonomous vehicles, anomalies can range from sensor failures due to bad lighting conditions, over pedestrians suddenly crossing the street, to abnormal driving patterns from other vehicles. 
The different contexts in which anomalies arise, and their inherent unpredictability, pose a challenge to their detection. Many recent approaches focus on unsupervised learning of models describing normality. The idea is to detect anomalies by their deviation from the learned model of normality~\cite{hasanLearningTemporalRegularity2016,liuFutureFramePrediction2018, moraisLearningRegularitySkeleton2019, termohlenCornerCaseDetection2019, parkLearningMemoryGuidedNormality2020}. Typically, the learning of normality is achieved by exploiting that models learn their training data's underlying patterns, which implicitly makes them learn a notion of normality corresponding to their training data~\cite{parkLearningMemoryGuidedNormality2020}.

At the same time, world models emerged in the field of \ac{RL} as a way to enable agents to predict the future conditioned on actions. This led to outstanding results in sparse reward and complex control tasks. Based on these successes, the question arises whether such world models, which learn a compact representation of the world and predict future changes, can be used outside of RL.

\begin{figure}[t]
\includegraphics[width=1\columnwidth]{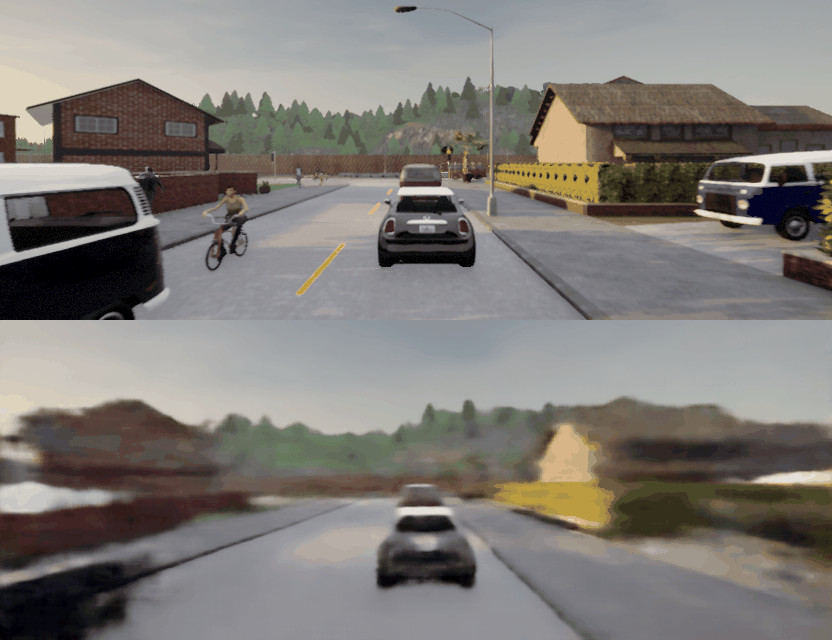}
\centering
\caption{The bottom row shows a scene reconstruction of a world model~\cite{huModelBasedImitationLearning2022} Compared to the ground truth, the model cannot recover all scene components, such as the bicyclist. This phenomenon can be exploited through a clear definition of normality and targeted training to detect anomalies.}
\label{fig:worldmodel_reco_bicyclist}
\end{figure}

In this work, we introduce anomaly detection methods for autonomous driving in Sec.~\ref{sec:ADonVD} and world models in Sec.~\ref{sec:world_models}. In Sec.~\ref{sec:anomaly_detection_with_world_models}, we describe our concept for the detection of anomalies with world models and conclude our thoughts in Sec.~\ref{sec:conclusion}.

\section{Anomaly Detection in Autonomous Driving}
\label{sec:ADonVD}

\definecolor{light-green}{RGB}{213,245,227}

\begin{figure*}[ht!]
\centering
\begin{tikzpicture}[node distance=0.9cm]

    \node[draw, circle, inner sep=0pt, minimum size=0.8cm, xshift=-4cm, fill=gray!25] (z0) {$s_{t-2}$};
    \node[draw, circle, inner sep=0pt, minimum size=0.8cm, above of=z0, xshift=0cm, yshift=0.3cm, fill=light-green] (a0) {$a_{t-2}$};

    \node[left of=z0] (dots_l) {$\ldots$};

    \node[draw, circle, inner sep=0pt, minimum size=0.8cm, xshift=0cm, fill=gray!25] (z1) {$s_{t-1}$};
    \node[draw, circle, inner sep=0pt, minimum size=0.8cm, below of=z1, xshift=0
    cm, yshift=-2.5cm, fill=light-green] (o1) {$o_{t-1}$};
    \node[draw, circle, inner sep=0pt, minimum size=0.8cm, below of=z1, xshift=0cm, yshift=-1.3cm] (oo1) {$\hat{o}_{t-1}$};
    \node[draw, circle, inner sep=0pt, minimum size=0.8cm, above of=z1, xshift=0cm, yshift=0.3cm, fill=light-green] (a1) {$a_{t-1}$};

    \coordinate (mid_z) at ($(z0)!0.5!(z1)$);
    \node[draw, rectangle, rounded corners, minimum width=2cm, minimum height=1cm] (m0) at (mid_z) {$Transition$};

    \node[draw=none, below of=m0, xshift=-0.4cm, yshift=-1.2cm] (embedding_label) {$Embedding$};
    
    \coordinate (mid_enc1) at ($(o1)!0.52!(m0.south)$);
    \path let \p1 = ($(m0.south)-(o1)$), \n1 = {atan2(\y1,\x1)} in
    \pgfextra{\xdef\angle{\n1}};
    \draw[name path=encoder1, rotate=(270+\angle), rounded corners=2.25pt] ($(mid_enc1)+(-0.4,-0.4)$) -- ($(mid_enc1)+(0.4,-0.4)$) -- ($(mid_enc1)+(0.15,0.4)$) -- ($(mid_enc1)+(-0.15,0.4)$) -- cycle;

    \coordinate (mid_dec1) at ($(oo1)!0.52!(z1)$);
    \path let \p1 = ($(z1)-(oo1)$), \n1 = {atan2(\y1,\x1)} in
    \pgfextra{\xdef\angle{\n1}};
    \draw[name path=decoder1, rotate=(270+\angle), rounded corners=2.25pt] ($(mid_dec1)+(-0.4,-0.4)$) -- ($(mid_dec1)+(0.4,-0.4)$) -- ($(mid_dec1)+(0.15,0.4)$) -- ($(mid_dec1)+(-0.15,0.4)$) -- cycle;


    \path[name path=o1path] (o1) -- (mid_enc1);
    \draw[->, name intersections={of=o1path and encoder1}] (o1) -- (intersection-1);
    \path[name path=z1path] (mid_enc1) -- (m0.south);
    \draw[->, name intersections={of=z1path and encoder1}] (intersection-1) -- (m0.south);

    \path[name path=o1path] (z1) -- (mid_dec1);
    \draw[->, name intersections={of=o1path and decoder1}] (z1) -- (intersection-1);
    \path[name path=z1path] (mid_dec1) -- (oo1);
    \draw[->, name intersections={of=z1path and decoder1}] (intersection-1) -- (oo1);

    \coordinate (mid_emb1) at ($(mid_enc1)!0.5!(mid_dec1)$);

    \node[draw, circle, inner sep=0pt, minimum size=0.8cm, right of=z1, xshift=3cm, fill=gray!25] (z2) {$s_{t}$};
    \node[draw, circle, inner sep=0pt, minimum size=0.8cm, below of=z2, xshift=0cm, yshift=-2.5cm, fill=light-green] (o2) {$o_{t}$};
    \node[draw, circle, inner sep=0pt, minimum size=0.8cm, below of=z2, xshift=0cm, yshift=-1.3cm] (oo2) {$\hat{o}_{t}$};
    \node[draw, circle, inner sep=0pt, minimum size=0.8cm, above of=z2, xshift=0cm, yshift=0.3cm, fill=light-green] (a2) {$a_{t}$};

    \coordinate (mid_z) at ($(z1)!0.5!(z2)$);
    \node[draw, rectangle, rounded corners, minimum width=2cm, minimum height=1cm] (m1) at (mid_z) {$Transition$};
    
    \coordinate (mid_enc2) at ($(o2)!0.5!(m1.south)$);
    Calc Angle
    \path let \p1 = ($(m1.south)-(o2)$), \n1 = {atan2(\y1,\x1)} in
    \pgfextra{\xdef\angle{\n1}};
    Draw centered and rotated trapezoid
    \draw[name path=encoder2, rotate=(270+\angle), rounded corners=2.25pt] ($(mid_enc2)+(-0.4,-0.4)$) -- ($(mid_enc2)+(0.4,-0.4)$) -- ($(mid_enc2)+(0.15,0.4)$) -- ($(mid_enc2)+(-0.15,0.4)$) -- cycle;

    \coordinate (mid_dec2) at ($(oo2)!0.5!(z2)$);
    \path let \p1 = ($(z2)-(oo2)$), \n1 = {atan2(\y1,\x1)} in
    \pgfextra{\xdef\angle{\n1}};
    \draw[name path=decoder2, rotate=(270+\angle), rounded corners=2.25pt] ($(mid_dec2)+(-0.4,-0.4)$) -- ($(mid_dec2)+(0.4,-0.4)$) -- ($(mid_dec2)+(0.15,0.4)$) -- ($(mid_dec2)+(-0.15,0.4)$) -- cycle;

    \path[name path=o2path] (o2) -- (mid_enc2);
    \draw[->, name intersections={of=o2path and encoder2}] (o2) -- (intersection-1);
    \path[name path=z2path] (mid_enc2) -- (m1.south);
    \draw[->, name intersections={of=z2path and encoder2}] (intersection-1) -- (m1.south);

    \path[name path=o2path] (z2) -- (mid_dec2);
    \draw[->, name intersections={of=o2path and decoder2}] (z2) -- (intersection-1);
    \path[name path=z2path] (mid_dec2) -- (oo2);
    \draw[->, name intersections={of=z2path and decoder2}] (intersection-1) -- (oo2);

    \node[draw, circle, inner sep=0pt, minimum size=0.8cm, right of=z2, xshift=3cm, fill=gray!25] (z3) {${s}_{t+1}$};
    \node[draw, circle, inner sep=0pt, minimum size=0.8cm, below of=z3, xshift=0cm, yshift=-1.3cm] (oo3) {$\hat{o}_{t+1}$};
    \node[draw, circle, inner sep=0pt, minimum size=0.8cm, above of=z3, xshift=0cm, yshift=0.3cm, fill=light-green] (a3) {$a_{t+1}$};

    \coordinate (mid_dec3) at ($(oo3)!0.5!(z3)$);
    \path let \p1 = ($(z3)-(oo3)$), \n1 = {atan2(\y1,\x1)} in
    \pgfextra{\xdef\angle{\n1}};
    \draw[name path=decoder3, rotate=(270+\angle), rounded corners=2.25pt] ($(mid_dec3)+(-0.4,-0.4)$) -- ($(mid_dec3)+(0.4,-0.4)$) -- ($(mid_dec3)+(0.15,0.4)$) -- ($(mid_dec3)+(-0.15,0.4)$) -- cycle;

    \path[name path=o2path] (z3) -- (mid_dec3);
    \draw[->, name intersections={of=o2path and decoder3}] (z3) -- (intersection-1);
    \path[name path=z2path] (mid_dec3) -- (oo3);
    \draw[->, name intersections={of=z2path and decoder3}] (intersection-1) -- (oo3);

     \node[draw, circle, inner sep=0pt, minimum size=0.8cm, right of=z3, xshift=3cm, fill=gray!25] (z4) {${s}_{t+2}$};
     \node[draw, circle, inner sep=0pt, minimum size=0.8cm, below of=z4, xshift=0cm, yshift=-1.3cm] (oo4) {$\hat{o}_{t+2}$};

    \coordinate (mid_dec4) at ($(oo4)!0.5!(z4)$);
    \path let \p1 = ($(z4)-(oo4)$), \n1 = {atan2(\y1,\x1)} in
    \pgfextra{\xdef\angle{\n1}};
    \draw[name path=decoder4, rotate=(270+\angle), rounded corners=2.25pt] ($(mid_dec4)+(-0.4,-0.4)$) -- ($(mid_dec4)+(0.4,-0.4)$) -- ($(mid_dec4)+(0.15,0.4)$) -- ($(mid_dec4)+(-0.15,0.4)$) -- cycle;

    \path[name path=o2path] (z4) -- (mid_dec4);
    \draw[->, name intersections={of=o2path and decoder4}] (z4) -- (intersection-1);
    \path[name path=z2path] (mid_dec4) -- (oo4);
    \draw[->, name intersections={of=z2path and decoder4}] (intersection-1) -- (oo4);

    \coordinate (mid_z) at ($(z2)!0.5!(z3)$);
    \node[draw, rectangle, rounded corners, minimum width=2cm, minimum height=1cm] (m2) at (mid_z) {$Transition$};

    \coordinate (mid_z) at ($(z3)!0.5!(z4)$);
    \node[draw, rectangle, rounded corners, minimum width=2cm, minimum height=1cm] (m3) at (mid_z) {$Transition$};

    \node[draw=none, below of=m3, xshift=-3.2cm, yshift=-0.2cm] (embedding_label) {$Decoding$};

    \draw[->] (z0) -- (m0);
    \draw[->] (a0) -- (m0.north);
    \draw[->] (m0) -- (z1);
    \draw[->] (z1) -- (m1);
    \draw[->] (a1) -- (m1.north);
    \draw[->] (m1) -- (z2);
    \draw[->] (z2) -- (m2);
    \draw[->] (a2) -- (m2.north);
    \draw[->] (m2) -- (z3);
    \draw[->] (z3) -- (m3);
    \draw[->] (a3) -- (m3.north);
    \draw[->] (m3) -- (z4);

    \node[right of=z4] (dots_r) {$\ldots$};

\end{tikzpicture}
\label{fig:world_model}
\caption{A world model during inference, given high dimensional observations $o_t\ldots o_{t-i} \in \Omega$ and past and planned actions $a_{t-i}\ldots a_{t+j} \in A$ at time $t$. All state \textbf{transitions} up to $s_t \in S$ are computed with a \textit{representation model} $p(s_t \mid s_{t-1},a_{t-1},o_t)$, where the observations are first being \textbf{embedded}. The embedding of actions is possible but optional. Future state \textbf{transitions} can be computed with the \textit{prediction model} $p(s_t \mid s_{t-1},a_{t-1})$ based on the Markov assumption, where each state only depends on its predecessor. With the \textit{observation model} $p(o_t \mid s_t)$, reconstructions $\hat o_t \in \Omega$ can be \textbf{decoded} from state $s_t$.}
\end{figure*}
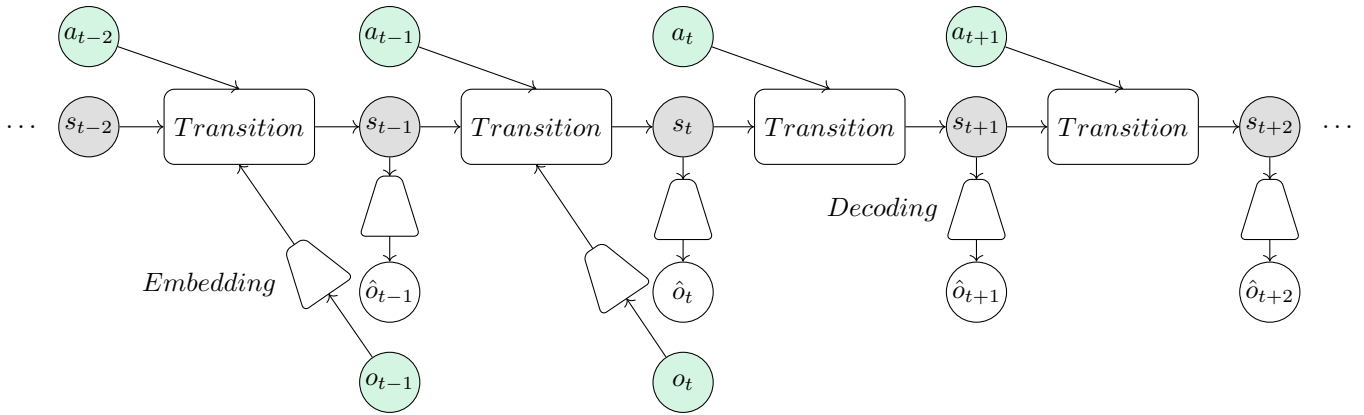

There are endless possibilities of deviations from normality, which makes it hard to formally define anomalies. Termöhlen et al.~\cite{termohlenCornerCaseDetection2019} were the first to formulate a definition in the domain of autonomous driving: “A corner case is given, if there is a non-predictable relevant object/class in [a] relevant location”. This section provides an overview of works on defining and detecting anomalies.

\textbf{Anomaly Types.} Breitenstein et al.~\cite{breitensteinSystematizationCornerCases2020} developed a five-level systematization of corner cases, focussing on camera data. Heidecker et al. generalized the work and extended it by a method layer, which describes corner cases “due to uncertainty inherent in the
methodology or the data”~\cite{heidecker_application}. The first three layers describe anomalies in the surrounding environment. The fourth layer describes anomalies that stem from the utilized software stack itself. The \textit{sensor layer} encompasses anomalies introduced by sensor attributes. The \textit{content layer} describes anomalies detectable within a single frame, such as changing weather conditions or unknown objects. The \textit{temporal layer} includes anomalies that become apparent only when analyzing multiple frames focusing on behavior.

The \textit{method layer} describes anomalies introduced by the used methodology, including the underlying training data. These may be expressed through uncertainty or other means. Unlike the previous layers, the method layer defines anomalies depending on the system's capabilities. Zhou and Beyerer argue similarly, defining corner cases as “interpretation problem[s] in the networks” based on training and test data~\cite{Zhou_Corner_2023_IV}.

\textbf{Detection Methods.} As one of the early works,
Breitenstein et al.~\cite{breitensteinCornerCasesVisual2021} investigated concepts for detecting anomalies for autonomous driving based on camera data, classifying approaches into reconstructive, generative, predictive, confidence score, and feature extraction-based methods. Bogdoll et al. extended the work by including lidar- and radar-based methods~\cite{bogdollAnomalyDetectionAutonomous2022}. Here, we provide an overview of these. 

\textit{Reconstructive approaches} aim to detect anomalies by reconstructing input frames from a compressed representation. Anomalies are identified when the reconstruction error or another suitable metric exceeds a predefined threshold, indicating the model's inability to represent the input accurately. Reconstruction-based methods build on the assumption that reconstructive models trained on normal input data fail to reconstruct anomalies. Similarly, \textit{Generative methods} are also based on reconstructing methods, but “also regard the discriminator’s decision or the distance between the generated and the training distribution”~\cite{breitensteinCornerCasesVisual2021}.

\textit{Predictive methods} predict future frames based on preceding frames. Subsequently, anomalies are detected when a significant deviation between the prediction and the actual observation is encountered. The deviation can be measured with techniques from reconstruction-based approaches, e.g., reconstruction error.

\textit{Confidence score-based methods} involve estimating the uncertainty associated with a model's prediction. Anomalies are detected when the confidence score of a model is low. Suggesting that the model is uncertain about the prediction might indicate that the observed data is potentially abnormal.

Lastly, \textit{feature extraction based methods} detect anomalies by transforming the input data into a lower-dimensional feature space, which can emphasize meaningful patterns and separate anomalous data points from normal ones. Techniques such as deep feature extraction, clustering algorithms, and one-class support vector machines can be employed in this context. Contrary to previous methods, this approach only allows classification, not pixel or point-wise detection.

For training and evaluation of such such methods, various datasets are available~\cite{Bogdoll_Addatasets_2022_VEHITS, Bogdoll_Impact_2023_ICCRE, Bogdoll_Perception_2023_IV}. Oftentimes, normality is represented by the classes and data provided by Cityscapes~\cite{Cordts2016TheCD}, while anomalies include categories such as costumed people, lost cargo, or animals.
\section{World Models}
\label{sec:world_models}
This section provides a high-level characterization of world models as well as important examples from the literature. World models originate from the field of Reinforcement Learning. Here, for every state transition, an agent additionally receives a reward. Given the goal to maximize the expected cumulative reward, the question arises: Which actions to take? In model-free RL, the agent directly optimizes for this target based on its interactions with the environment. In model-based Reinforcement Learning, however, a dynamics model is learned first in order to predict the dynamics of the environment. This way, the costly policy learning can be done purely in the dynamics model.

While the term \textit{world model} for such a dynamics model dates back to the early days of Artificial Intelligence~\cite{Hu_Neural_2022_PhD}, in modern days Ha and Schmidhuber coined the term~\cite{haWorldModels2018}. However, they did not provide a clear definition. They were the first who modeled “dynamics observed from high dimensional visual data where [the] input is a sequence of raw pixel frames”~\cite{haWorldModels2018}. LeCun broadly defines a world model as an “internal model of how the world works”~\cite{lecun}. Following Kendall, “A world model is a generative model that is able to predict the next state conditioned on an action”~\cite{Kendall_E2EAD_2023_CVPR}. Chen et al. describe world models based on their capabilities to provide “abstract perceptual representations” and “explicit future predictions”~\cite{chen2023endtoend}. Combining these concepts, we define a world model as follows:
\begin{displayquote}
\textit{A world model embeds sensory observations into a latent state, predicts action-conditioned state transitions, and is able to decode into observation space.}
\end{displayquote}

Such a stochastic, generative \textit{world model} $\mathcal{W}$, as shown in Figure~\ref{fig:world_model}, can be described by three conditional probability distributions~\cite{hafnerLearningLatentDynamics2019,hafnerDreamControlLearning2020}, where the \textit{representation model}
\begin{equation}
p(s_{t} \mid s_{t-1},a_{t-1},o_t)
\end{equation}
describes the dependence of a latent state on the associated observation, the \textit{prediction model} 
\begin{equation}
p(s_t \mid s_{t-1},a_{t-1})
\end{equation}
describes the transition from one latent state to the next. It is possible to predict multiple next states by sampling: $s_{t+1}\sim p(s_{t+1} \mid s_t,a_t)$. Finally, the \textit{observation model}
\begin{equation}
p(o_{t} \mid s_t)
\end{equation}
describes the dependence of an observation on the associated latent state, which allows sampling reconstructions from it: $\hat o_t\sim p(o_{t} \mid s_t)$. In addition, a world model can have further connections or heads, e.g., an additional head for rewards is typical in Reinforcement Learning. Since a world model consists of multiple deep neural networks (DNN), Ha and Schmidhuber proposed a separation into a \textit{Vision} and a \textit{Memory} component~\cite{haWorldModels2018}. Their vision model is responsible for compressing observations to compact representations, while the memory model is responsible for predicting latent states. As shown in Figure~\ref{fig:world_model}, we find naming those components \textit{Embedding} and \textit{Transition} are more concise choices. Additionally, we explicitly introduce the \textit{Decoding} model to emphasize the property of transforming abstract representations into explicit reconstructions, which aligns well with the three conditional probability distributions.

\subsection{Problem Formulation}

To model the interaction of an \textbf{agent} and a realistic \textbf{environment}, we adapt the well-known terminology from \ac{POMDP} and assume Markovian properties, as shown in Figure~\ref{fig:terminology}. It can be assumed that no \textbf{actor} within the environment has access to its latent \textbf{state}. In the domain of autonomous driving, any agent controlling an actor thus relies on sensory \textbf{observations} from the actor itself. The agent can be understood as a software component that controls the vehicle, given its sensor data as observations. With each \textbf{action}, the internal state of the environment gets updated.

\tikzset{
  frame/.style={
    rectangle, draw,
    text width=6em, text centered,
    minimum height=2em,fill=white,
    rounded corners,
  },
  env/.style={
    rectangle, draw,
    text width=10em, text centered, anchor=north, text depth = 4.5em,
    minimum height=6em,
    rounded corners,
  },
  line/.style={
    draw, -{Latex},rounded corners=3mm,
  }
}
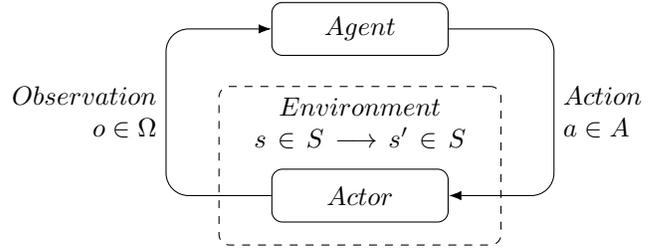
\begin{figure}[ht]
\centering
\begin{tikzpicture}[node distance = 4cm]
\node [frame] (agent) {$Agent$};
\node [env, dashed, below=0.4cm of agent] (environment) {$Environment$\\ $s\in S \longrightarrow s' \in S$};
\node [frame, below=1.5cm of agent] (actor) {$Actor$};
\draw[line] (agent.east) -- ++ (1.4,0) |- (actor.east) 
node[right,pos=0.25,align=left] {$Action$\\ $a \in A$};
\coordinate[left=12mm of actor] (P);
\pgfmathsetmacro{\Ldist}{4mm}
\draw[line] (actor.west) -- ++ (-1.4,0) |- 
(agent.west) node[left, pos=0.25, align=right] {$Observation$\\ $o \in \Omega$};
\end{tikzpicture}
\caption{Interaction of an agent with an actor in an environment, where $A$ is a set of actions, $S$ is a set of states, and $\Omega$ is a set of observations. Given an observation $o$, the action $a$ of the agent results in a state transition $s \longrightarrow s'$ of the environment.}
\label{fig:terminology}
\end{figure}

\subsection{Embedding Models}
While contrastive approaches exist~\cite{hafnerDreamControlLearning2020}, most recent works implement VAEs as the embedding model to embed observations in a latent state space and utilize a reconstruction loss during training~\cite{watterEmbedControlLocally2015, haWorldModels2018, hafnerLearningLatentDynamics2019, hafnerDreamControlLearning2020, beckerUncertaintyDeepState2022, hafnerMasteringAtariDiscrete2022, hafnerMasteringDiverseDomains2023}. This also allows for decoding latent representations. Given a latent state $s_t$, the corresponding decoder can be used to obtain a reconstruction $\hat o_t$ of the original observation $o_t$. The state space structure is implicitly given by the embedding model.
A common choice are Gaussian state spaces. Ha \& Schmidhuber~\cite{haWorldModels2018} use a convolutional \ac{VAE} to encode two-dimensional RGB image frames into low dimensional vectors $\mu, \sigma \in \mathbb{R}^{32}$, which represent the parameters of a Gaussian distribution $\mathcal{N}(\mu,\sigma I)$. Latent states $s \in \mathbb{R}^{32}$ are sampled from this distribution. The authors claim that the Gaussian prior makes the world model “more robust to unrealistic [$s$] vectors” predicted by their transition model. The locally linear latent state space dynamics model Embed to Control (E2C)~\cite{watterEmbedControlLocally2015} uses a similar VAE-type encoder, also resulting in a Gaussian state space.
PlaNet, a model-based agent proposed by Hafner et al.~\cite{hafnerLearningLatentDynamics2019}, performs planning over imagined latent state trajectories, using the embedding model from Ha and Schmidhuber~\cite{haWorldModels2018}. The reinforcement learning agent Dreamer, proposed by Hafner et al.~\cite{hafnerDreamControlLearning2020}, utilizes the previously introduced world model PlaNet. In contrast, Dreamers recent iterations DreamerV2~\cite{hafnerMasteringAtariDiscrete2022} and DreamerV3 \cite{hafnerMasteringDiverseDomains2023} replace the Gaussian state space with a categorical state space, where image frames are mapped to a categorical distribution. From this distribution, a latent state $s$ encoded in categoricals is sampled.

\ac{vrkn}, proposed by Becker \& Neumann~\cite{beckerUncertaintyDeepState2022}, encode observations using a Neural Network with two output heads, which directly maps an observation to an intermediate representation $w_t$ and its diagonal covariance $\sigma^{w_t}$. The latent state is then calculated using Bayes rule for Gaussian distributions (Kalman updates).

Hu et al. presented their \ac{MILE} for predictions in urban environments~\cite{huModelBasedImitationLearning2022, Hu_Neural_2022_PhD}. Compared to previous works, which apply world models in the context of \ac{RL}, they use \ac{IL} to train their model. Inspired by Philion and Fidler~\cite{philion2020lift}, they encode high-resolution observations and lift the resulting features into 3D space based on a learned depth probability distribution for each feature. A pooling operation into \ac{BeV} space and mapping to a 1D vector follow.

\subsection{Transition Models}
A central idea of a world model is to be able to look ahead by predicting future latent states. Given a latent state $s_t$ and an action $a_t$, a world model is able to predict future states. Therefore, in addition to an embedding model as described above, a \textit{transition model} is needed. 

Inspired by Stochastic Optimal Control algorithms, E2C~\cite{watterEmbedControlLocally2015} assumes locally linear dynamics at each time step to approximate global non-linear dynamics. Future latent states are sampled from a Gaussian distribution $\mathcal{N}(A_t \mu_t + B_t a_t + o_t, C_t)$, where $\mu_t$ and $\sigma_t$ are the parameters of $s_t$ computed by the embedding model. $A(s_t)$ is the local Jacobian with respect to $s_t$,  $B(s_t)$ the local Jacobian with respect to $a_t$, $o(s_t)$ is an offset and $C_t = A_t \Sigma_t {A}^T_t + H_t$ with system noise $H_t$.

Ha \& Schmidhuber~\cite{haWorldModels2018} use a Recurrent Neural Network (RNN) with a Mixture Density Network (MDN) output layer, called MDN-RNN, to model the probability distribution $p(s_{t+1} \mid s_t, a_t, h_t)$, from which $s_{t+1}$ can be sampled. The additional parameter $h_t$ is the RNNs hidden state.

PlaNet~\cite{hafnerLearningLatentDynamics2019} uses a Recurrent State Space Model (RSSM)~\cite{hafnerLearningLatentDynamics2019} to model latent dynamics. RSSMs combine State Space Models (SSM) and Recurrent Neural Networks by introducing a deterministic path to SSMs, consisting of the hidden states of an RNN. More precisely, the transition model is an RSSM consisting of a deterministic state model $h_t = f(h_{t-1}, s_{t-1}, a_{t-1})$ implemented as an RNN, and a stochastic state model $s_t \sim p(s_t\ |\ h_t)$.
The sequence of hidden states $h_{1:T}$ is called the deterministic path, while the sequence of stochastic latent states $s_{1:T}$ is called the stochastic path of the RSSM.
While purely deterministic transitions ``prevent the model from capturing multiple futures'', the stochastic transitions in state space models make “it difficult to remember information over multiple time steps”~\cite{hafnerLearningLatentDynamics2019}. The RSSM of PlaNet is reused in all iterations of Dreamer.

VRKN~\cite{beckerUncertaintyDeepState2022} improve on RSSMs by eliminating the need for a deterministic path with a more principled modeling of aleatoric and epistemic uncertainty. This is achieved by using an inference scheme consistent with the generative model, i.e., by imposing the same assumptions on World Model learning and inference. Latent state estimates are forwarded using closed-form Gaussian marginalization, and then updated based on the respective observation using Kalman updates, thereby incorporating information from previous states. Epistemic uncertainty is captured using Monte Carlo dropout layers. Aleatoric uncertainty is captured in two ways: First, the latent state is a distribution capturing multiple possible futures, which provides information about aleatoric uncertainty. Second, Kalman updates enable the update of state estimates based on the respective observation. RSSMs can not correct latent state estimates and thus explain differences between estimates and observations implicitly by the transition model. VRKN can correct differences between state estimates and observations with Kalman updates. This eliminates the overestimation of aleatoric uncertainty present in RSSMs.

\ac{MILE}~\cite{huModelBasedImitationLearning2022} follows a similar approach as Hafner et al~\cite{hafnerLearningLatentDynamics2019}. The difference is that the deterministic history $h_t$ in MILE is obtained only from the previous history $h_{t-1}$ and stochastic state $s_{t-1}$, i.e., $h_t = f(h_{t-1}, s_{t-1})$ where $f$ is modeled as an RNN. Actions are later introduced in the stochastic state model $s_t \sim p(s_t\ |\ h_t, a_{t-1})$.

In addition to this academic progress, world models are also finding their way into the industry. GAIA-1 by Wayve, trained on fleet data, “leverages video, text, and action inputs to generate realistic driving videos”~\cite{Wayve_GAIA_2023_Web}. Similarly, Elluswamy presented early results of a “general world model” developed by Tesla, also trained on fleet data~\cite{Elluswamy_WAD_2023_CVPR}.

\section{Anomaly Detection with World Models}
\label{sec:anomaly_detection_with_world_models}

After presenting anomalies in the domain of autonomous driving and recent world models, we now introduce our concept of utilizing world models to detect multiple forms of anomalies. Following the approach presented by Breitenstein et al.~\cite{breitensteinCornerCasesVisual2021}, we map the previously introduced detection methods to corner case levels~\cite{breitensteinSystematizationCornerCases2020, heidecker_application}, as shown in Table~\ref{tab:method_cc}. We mark a method only as applicable if it is the primary detection approach, since oftentimes multiple methods are being used, supplementing each other. While \textit{sensor layer} corner cases can affect sensory observations and might thus be detectable, we focus on anomalies in the surrounding environment. Thus we deem anomalies on the \textit{sensor layer} out of scope.

\begin{table}[t]
\caption{Applicability of anomaly detection approaches with world models for corner case levels}
\label{tab:method_cc}
\resizebox{\columnwidth}{!}{%
\begin{tabular}{@{}lcccccc@{}}
\toprule
\textbf{Layer}              & \multicolumn{2}{c}{Sensor} & \multicolumn{3}{c}{Content} & Temporal \\ \midrule
\textbf{Level}              & Hardware     & Physical    & Domain   & Object  & Scene  & Scenario \\ \midrule
\textbf{Reconstructive}     & ---            & ---           & \cmark        & \cmark       & \cmark      & ---        \\
\textbf{Generative}         & ---            & ---           & \cmark        & \cmark       & \cmark      & ---        \\
\textbf{Predictive}         & ---            & ---           & ---        & ---       & ---     & \cmark        \\
\textbf{Confidence Score}   & ---            & ---           & \cmark        & ---       & ---      & ---
        \\
\textbf{Feature Extraction} & ---            & ---           & \cmark        & ---       & ---      & ---
       
\end{tabular}%
}
\end{table}

World models are especially interesting for anomaly detection, as they are able to perform all detection approaches from the literature: As the \textit{embedding model} is typically implemented as a stochastic \ac{VAE}, it has both reconstructive and generative capabilities. The core purpose of the \textit{transition model} is prediction. For both the embedding model and the transition model, epistemic uncertainty estimates are possible. Finally, since we have access to the model, features are available. This allows world models to detect a wide variety of anomalies in a single pass, compared to a plethora of models which would be necessary otherwise. In addition, the end-to-end training approach of world models allows for the manifestation of a common definition of normality.

In general, the field of anomaly detection, especially for autonomous driving, faces several challenges at the moment~\cite{bogdoll_description, bogdollAnomalyDetectionAutonomous2022, Bogdoll_Perception_2023_IV}. We want to address them systematically, introducing several assumptions in the following.

\textbf{Normality.}
Since we aim to detect anomalies based on the capabilities of a world model, the learned normality of this world model is of utmost importance. However, real-world data collected at scale will contain unlabeled anomalies~\cite{Du_Unknown_2022_CVPR}. Thus, achieving a clear disjunction between normality and anomalies, represented by training and evaluation data, is hard. This is especially important for un- and self-supervised approaches, which learn from all patterns included in the training data. If they pick up unknown anomalies from the training data, which are defined as anomalies in the evaluation data, this can lead to a situation where these human-defined anomalies are no longer anomalies for the detecting system.

\textbf{Mapping.}
Typically, a detection system provides “metric-based assessments of situations”~\cite{bogdoll_description}. To provide more context, it is often of interest to map detections to a corner case category, as shown in Table~\ref{tab:method_cc}. While the Table only shows the applicability of methods, the actual assignment during inference is challenging. For example, if a latent representation of an input deviates from learned representations of normal samples, a domain shift or a sufficiently large unknown object could be the reason.

\textbf{Evaluation.}
For evaluation, concrete scenarios containing anomalies are necessary. However, deriving concrete scenarios from the rather broad corner case categories in a scalable fashion is hard~\cite{bogdoll_description, Bogdoll_Ontology_2022_ECCV}. In most benchmarks, the anomaly type is defined as either unknown objects, domain shifts, or abnormal behavior~\cite{Bogdoll_Perception_2023_IV}. Accordingly, the computer vision community focuses on “contextual anomalies on the scene level”~\cite{bogdollAnomalyDetectionAutonomous2022}. Addressing these challenges, we define the following assumptions:

\begin{itemize}
    \item To properly define anomalies, full control over both the training and evaluation data is necessary
    \item For a precise definition of normality, anomalies cannot be part of the training data
    \item For the evaluation, human-defined anomalies are required, which must be in agreement with the normality defined by the training data
\end{itemize}

In order to have full control over both training and test data, a controlled environment is necessary. The next sections will first introduce training and test data, followed by the general inference concept of the model and concrete methods for the detection of anomalies. 

\subsection{Training Data}
As described earlier, the definition of an anomaly depends on the capabilities of the perception system, which depends on the data it was trained on. Following this concept, in order to detect anomalies, a concept of normality needs to be defined first. We can clearly define both the static and the dynamic environment of the training dataset~$\mathcal{D}_{norm}^{train}$ based on several attributes: \textit{Region, Weather, Time of Day, Objects, Actors, and Behaviors}. In a similar fashion, these can be found in the 6-Layer PEGASUS model~\cite{scholtes_pegasus}. Based on a simulation engine~\cite{Dosovitskiy17,muetsch2023modelbased}, these can be set in a reproducible and deterministic way. For the behavior of the ego-vehicle during training, a driving model providing expert demonstrations is well suited~\cite{huModelBasedImitationLearning2022,zhang2021roach,Dosovitskiy17}. Combining the static and dynamic attributes of the environment and a driving model for the ego vehicle allows for the assembly of a well-defined and large-scale dataset for training. If anomaly detection approaches shall be compared based on an evaluation dataset~$\mathcal{D}_{ano}^{eval}$, they need to share a common definition of normality in the form of a prescribed training dataset~$\mathcal{D}_{norm}^{train}$. Today, this is typically not the case, which makes it impossible to truly compare approaches~\cite{Bogdoll_Perception_2023_IV}. For training details, we refer to common literature in the domain of world models~\cite{haWorldModels2018, hafnerLearningLatentDynamics2019, hafnerDreamControlLearning2020, huModelBasedImitationLearning2022}. 

\subsection{Evaluation Data}

For the evaluation dataset~$\mathcal{D}_{ano}^{eval}$ we need to integrate anomalies purposefully. Since the training dataset is well defined, this is possible on all corner case levels separately: On the \textit{sensor layer}, perturbations can be added to the sensor configuration or sensor data. On the \textit{domain level}, different regions, weather conditions, or daytimes can be chosen. For both the \textit{object} and \textit{scene level}, unknown objects or known objects in atypical places can be inserted. Finally, for the \textit{scenario layer}, atypical behaviors can be defined or generated. For all of these anomalies, ground truth is available, which allows us to use them for evaluation. In addition, it is possible to combine corner case levels. 

\begin{figure}[tbp]
  \def\svgwidth{\columnwidth}
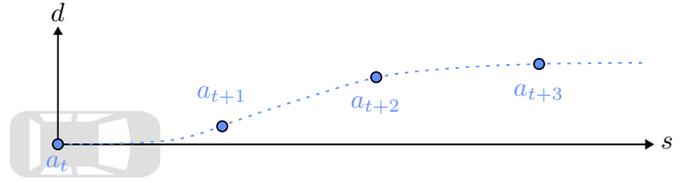
  \caption{Planned actions in the context of autonomous driving. Based on a vehicle model, the \textit{planning module} determines a finite list of actions for the ego vehicle in order to reach planned future vehicle states.}
  \label{fig:frenet}
\end{figure}

\begin{figure*}[t!]
\includegraphics[width=1\textwidth]{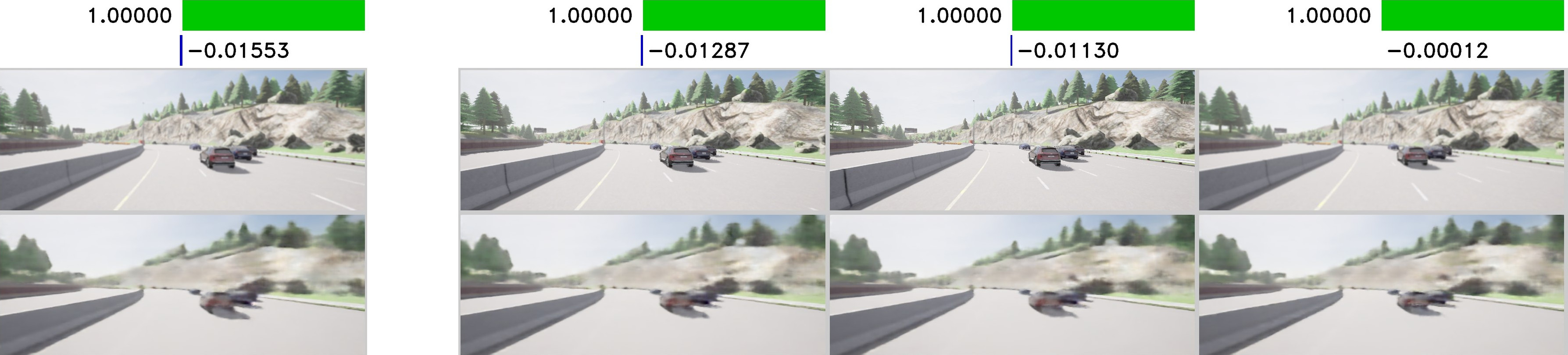}
\centering
\caption{On the left, the last input frame for the prediction and its reconstruction are shown. The bottom row on the right shows the predictions of a world model conditioned on actions, where each action consists of acceleration and steering angle values as shown in the top row~\cite{huModelBasedImitationLearning2022}. Compared with the ground truth, the model is able to predict normal behavior. Under the hypothesis that it cannot predict atypical behavior unseen during training, differences between future observations and the predictions can be used for anomaly detection.}
\label{fig:worldmodel_prediction}
\end{figure*}

\subsection{Model Inference}

As shown in Figure~\ref{fig:world_model}, future actions are necessary for the rollout of a world model. In Reinforcement Learning or Imitation Learning, typically a policy is learned to select an action $a_t=(acc_t,\delta_t)$ during inference, as visible in Figure~\ref{fig:worldmodel_prediction}. Applied to the domain of anomaly detection for autonomous driving though, a set of planned actions is known beforehand which can be utilized for the rollout, similar to the known actions used for planning with the PLaNet model~\cite{hafnerLearningLatentDynamics2019}. In autonomous driving, it is still common to divide the overarching task into a subset of modules, most importantly perception, prediction, planning, and control. Given information about the state of the actor and some vehicle model, the planning module is able to compute a list of planned actions, as shown in Figure~\ref{fig:frenet}. While the state of the actor is implicitly included in the environment state $s$, some attributes of it can be explicitly known or measured, as shown in Equation~\ref{eq:actorstate}.

\begin{equation}
\begin{aligned}
s_{\text{actor}} = \left\{
\begin{tabular}{ll}
\text{coordinates}: & $(x, y, z)$, \\
\text{rotation}: & $(\phi,\theta,\psi)$, \\
\text{velocity}: & $v$, \\
\text{acceleration}: & $acc$, \\
\text{steering\_angle}: & $\delta$, \\
\vdots \\
\end{tabular}
\right\}
\end{aligned}
\label{eq:actorstate}
\end{equation}

Now, given observations, a list of past and planned actions, and a world model $\mathcal{W}$, a sequence of future latent states, also referred to as a \textit{rollout trajectory}, can be predicted. Based on

\begin{equation}
p(o_{1:T} \mid a_{0:T-1})\triangleq \mathbf{E}_{p(s_{1:T} \mid a_{0:T-1})}\Biggl[ \prod_{t=1}^Tp(o_t \mid s_t) \Biggr] 
\label{eq:distribution}
\end{equation}
which describes the next $T$ observations conditioned on $T$ given actions~\cite{hafnerLearningLatentDynamics2019}, we can sample $N$ times from the distribution in order to derive multiple futures in observation space: 
\begin{equation}
\{\hat o_{1:T}^n\}_{n=1}^N \sim p(o_{1:T} \mid a_{0:T-1})
\label{eq:sampling}
\end{equation}

\subsection{Anomaly Detection}

To the best of our knowledge, no existing method for anomaly detection based on sensor data utilizes a world model as described in Section~\ref{sec:world_models}. 
The anomaly detection method proposed by Chakraborty et al.~\cite{chakrabortyStructuralAttentionBasedRecurrent2023} is the only approach known to us which utilizes embeddings, latent state transitions, and decodings to detect anomalies. However, their transition model is not action-conditioned and their input consists of map and trajectory data instead of sensor observations. In the following, we will introduce methods from the literature, which can also be implemented based on a world model, from the categories \textit{reconstructive, generative, predictive, confidence score, and feature extraction}, as shown in Table~\ref{tab:method_cc}.

\textbf{Reconstructive and Generative.}
Most world models leverage embedding models which make use of a reconstructive training objective~\cite{haWorldModels2018, hafnerLearningLatentDynamics2019, hafnerDreamControlLearning2020, hafnerMasteringAtariDiscrete2022, hafnerMasteringDiverseDomains2023}, which form a reconstructive element inside a world model applicable to anomaly detection. While there are many anomaly detection approaches that use different forms of reconstructions~\cite{dibiasePixelwiseAnomalyDetection2021, grcic2021dense,grcic2023advantages}, their concepts can be applied to world models. It is important to keep in mind that well-trained VAEs generalize well and are able to reconstruct the unseen~\cite{Bogdoll_Compressing_2021_NeurIPS}, which is why special care is necessary during training and the design of anomaly detection methods.

Confronted with a \textit{domain level} corner case, the reconstruction quality can be poor for the entire frame. This can be detected, especially in comparison to cases where only certain regions of the reconstruction have poor quality. This classification task can be performed based on methods such as reconstruction quality~\cite{vu_anomaly_2019}, reconstruction probability~\cite{munjal_implicit_2020}, or even combinations with extracted features~\cite{abati_latent_2019, wang_image_2020}. Some of these methods also provide pixel-wise anomaly scores, which can be used for \textit{object level} or \textit{scene level} corner cases, which are hard to distinguish from a detection point of view. For example, Vojir et al.~\cite{vojirRoadAnomalyDetection2021} aim to detect anomalies on roads using a reconstructive approach.
The idea is to reconstruct the surface of the road and the remaining environment from a latent representation in such a way that the road reconstruction shows minimal error while the reconstruction of the remaining environment shows maximal error. The resulting pixel-wise errors are combined with a semantic segmentation output by feeding both into the semantic coupling module, which outputs two maps, one for the road class and one for the anomaly class.

\textbf{Predictive.}
Predictive capabilities form the core of a world model, enabling the prediction of future latent states and, based on the observation model, the reconstruction into the observation space, as shown in Figure~\ref{fig:worldmodel_prediction}. Given such predictions, either the  epistemic uncertainty or the comparison to future observations, which are available after $\Delta t$, can be used to detect \textit{scenario level} corner cases. When multiple futures are being predicted, distance metrics in their latent representations can be used to detect the prediction which generally aligns best with the ground truth.

Liu et al.~\cite{liuFutureFramePrediction2018} were the first to propose an approach to video anomaly detection involving the prediction of a future image frame and comparing the predicted frame with the ground truth frame. The approach adopts U-Net~\cite{ronnebergerUNetConvolutionalNetworks2015} to predict future frames based on the preceding image sequence. Intensity and gradient losses are directly computed between predicted and ground truth frames. To compute an optical flow loss, the authors leverage Flownet~\cite{dosovitskiyFlowNetLearningOptical2015} to estimate the optical flow between the predicted or ground truth frames and their preceding frames. An optical flow constraint boosts anomaly detection performance as it imposes motion consistency for normal events through capturing temporal information. Based on the assumption that normal events are well predicted, the final anomaly score is based on the Peak Signal-to-Noise Ratio (PSNR) between predicted and ground truth frames. Similarly, Termöhlen et al.~\cite{termohlenCornerCaseDetection2019} use a convolutional Autoencoder to predict an image frame from a sequence of preceding frames, but in the domain of camera-based autonomous driving. An error map is built from pixel-wise prediction errors and weighted subsequently based on semantic information. 

Chakraborty et al.~\cite{chakrabortyStructuralAttentionBasedRecurrent2023} propose a Structural Attention-based Recurrent VAE (SABer-VAE) for detecting anomalies in vehicle trajectories. The environment is modeled as a road map consisting of equidistant nodes on each lane. Furthermore, vehicle positions are given in coordinates at each timestep. The authors detect anomalies in observed vehicle trajectories by predicting future vehicle states and calculating a prediction loss from which a final anomaly score is derived. 

Their \textit{embedding model} consists of two paths. The primary encoding path models vehicle-vehicle interactions with a self-attention module. It takes vehicle positions and their relative distance to each other as inputs and transforms them, analogously to~\cite{haWorldModels2018}, but using a different architecture, into the parameters $\mu$ and $\sigma$ of a Gaussian latent state distribution. The secondary encoding path models lane-vehicle interactions and consists only of an attention module to calculate embeddings. Their \textit{transition model} uses a stochastic Koopman operator, where two Neural Networks predict the Koopman matrices in order to predict the “one-step future states of vehicles”. 

\textbf{Confidence- and Feature-based.}
The process of embedding environment observations into a latent space is one form of feature extraction. Based on these features or uncertainties w.r.t. to latent states, a whole observation or state can be classified as an anomaly~\cite{klausAnomalyDetectionLatent2022}, enabling the detection of \textit{domain level} corner cases. In single cases, where a large part of an image consists of \textit{object level} or \textit{scene level} corner cases, these can also be detected, but this dependence does not allow for the general detection of these levels. For example, Norlander \& Sopasakis~\cite{norlanderLatentSpaceConditioning2019} propose the \ac{CL-VAE} to detect anomalies in class-labeled data. The authors train a VAE using a Gaussian mixture model such that for each class in the data set, a Gaussian prior is fitted in the latent space. The authors show that anomalies lie between the clusters, which can be visually confirmed. The favorable structure of the latent space can be used for anomaly detection by employing techniques such as Isolation Forests~\cite{liuIsolationForest2008}. Another approach for anomaly detection in surveillance videos is proposed by Park et al.~\cite{parkLearningMemoryGuidedNormality2020}, who propose a memory module, which represents a storage of latent representations of normal image frames. 
A U-Net~\cite{ronnebergerUNetConvolutionalNetworks2015} based encoder extracts query features from a video frame. 
The queries are fed into the memory model to update the latent memory items. To reconstruct image frames, the memory items are concatenated with the query features and fed into the decoder, which outputs a reconstructed frame. 
Since “prediction can be considered as a reconstruction of the future frame using previous ones”~\cite{parkLearningMemoryGuidedNormality2020}, the model can be adapted to predict the next frame with minimal effort using the same underlying architecture and loss functions.
For training, the authors extend the reconstruction loss by adding a feature compactness as well as a feature separateness loss to condition the latent memory and query space. During inference, an anomaly score is obtained by combining the Euclidean distance between the extracted query and the closest memory item with the PSNR between input and reconstruction.

\section{Conclusion}
\label{sec:conclusion}

We presented a detailed characterization of \textit{world models} together with anomaly detection approaches from the literature which build on ideas already present in world models or applicable to our characterization. We show that world models hold great potential for the task of  anomaly detection in the context of autonomous driving. Along with a clear definition of anomalies and the categorization of anomaly detection approaches into predictive, generative, confidence score, and feature extraction-based methods, we showed that world models can be used to implement existing ideas from current approaches in a unified anomaly detection framework.
\section{Acknowledgment}
\label{sec:acknowledgment}
This work results from the jbDATA project supported by the German Federal Ministry for Economic Affairs and Climate Action of Germany (BMWK) and the European Union, grant number 19A23003H.

{\small
\bibliographystyle{IEEEtran}
\bibliography{references}
}

\end{document}